**Article title**

# Cross-Session 3D LiDAR and Camera Fusion for Robust Localization of Unmanned Aerial Vehicles in GPS-Denied Environments

**Authors**

Cong Hoang Quach[1], Chi Thanh Vo[2], Dong LT. Tran[2], Truong Son Nguyen[5], Manh Duong Phung[3,4], Thuan Hoang Tran[2,*]

**Affiliations**

1 – School of Electrical and Data Engineering, University of Technology Sydney, New South Wales, Australia
2 – Institute of Aerospace Engineering and Technology, Duy Tan University, Da Nang, Vietnam
3 – College of Engineering and Computer Science, VinUniversity, Hanoi, Vietnam
4 – Smart Green Transformation Center (GREEN-X), VinUniversity, Hanoi, Vietnam
5 – Department of Artificial Intelligence and Robotics, Sejong University, Seoul, South Korea

**Corresponding author's email address**

tranthuanhoang@duytan.edu.vn

**Abstract**

Accurate localization of unmanned aerial vehicles (UAVs) is essential for applications such as structural health monitoring, especially in environments where Global Positioning System (GPS) signals are denied or unreliable, like indoor spaces, tunnels, urban canyons, or areas beneath large structures. To address this challenge, we propose Cross-Fusion, a novel method for real-time UAV localization that integrates data from a 3D Light Detection and Ranging (LiDAR) and a monocular camera. A key contribution is its cross-session fusion strategy, which integrates visual and geometric information collected from multiple agents during routine baseline surveys to improve localization consistency and map completeness. The system employs LiDAR-based odometry for motion tracking and image-based feature matching via a single red-green-blue (RGB) camera to correct drift and improve accuracy. Unlike visual-inertial systems, Cross-Fusion maintains a simple sensor setup and avoids the complexity of stereo or global shutter configurations. Experimental results demonstrate that Cross-Fusion achieves localization accuracy comparable to GPS-based methods and performs reliably in challenging feature-sparse environments.

## 1. Introduction

Localization is fundamental to stable flight, precise trajectory tracking, and safe navigation because it determines the position and orientation of an unmanned aerial vehicle (UAV). This information provides critical feedback to flight controllers and mission planners for attitude control, path execution, and obstacle avoidance. Although most UAVs currently rely on GPS for localization, satellite signals are often weak, degraded, or unavailable in many real-world operating conditions [1]. This limitation is particularly critical for UAV-based infrastructure inspection in GPS-denied environments, such as bridge undersides, tunnels, and dense urban structures, where satellite visibility is frequently blocked by surrounding geometry. These environments also pose additional perception challenges because they often combine limited visual texture, repetitive patterns, and complex 3D structures.

As the low-altitude economy continues to expand, UAVs are increasingly being deployed for routine structural health monitoring and infrastructure maintenance missions [2, 3]. In such applications, reliable navigation around complex GPS-denied infrastructure is essential rather than optional. Therefore, robust localization methods that can integrate complementary sensing information are needed to support safe and consistent UAV operation in these challenging environments.

Common sensors used in UAVs include cameras, LiDAR, and inertial measurement units (IMU) [4, 5]. Cameras provide high-resolution imagery and rich visual detail suitable for map-building and visual odometry. However, their performance degrades in poor lighting or feature-sparse environments. LiDAR is particularly useful in low-visibility conditions where cameras fail, as it measures direct range by scanning the environment with laser beams. Nevertheless, its effectiveness can be limited in geometrically flat environments such as bridge surfaces. IMU measure a UAV's motion at high update rates and function independently of environmental conditions, but they suffer from drift over time due to the accumulation of measurement errors. For robust and accurate localization, algorithms should fuse data from these complementary sensors.

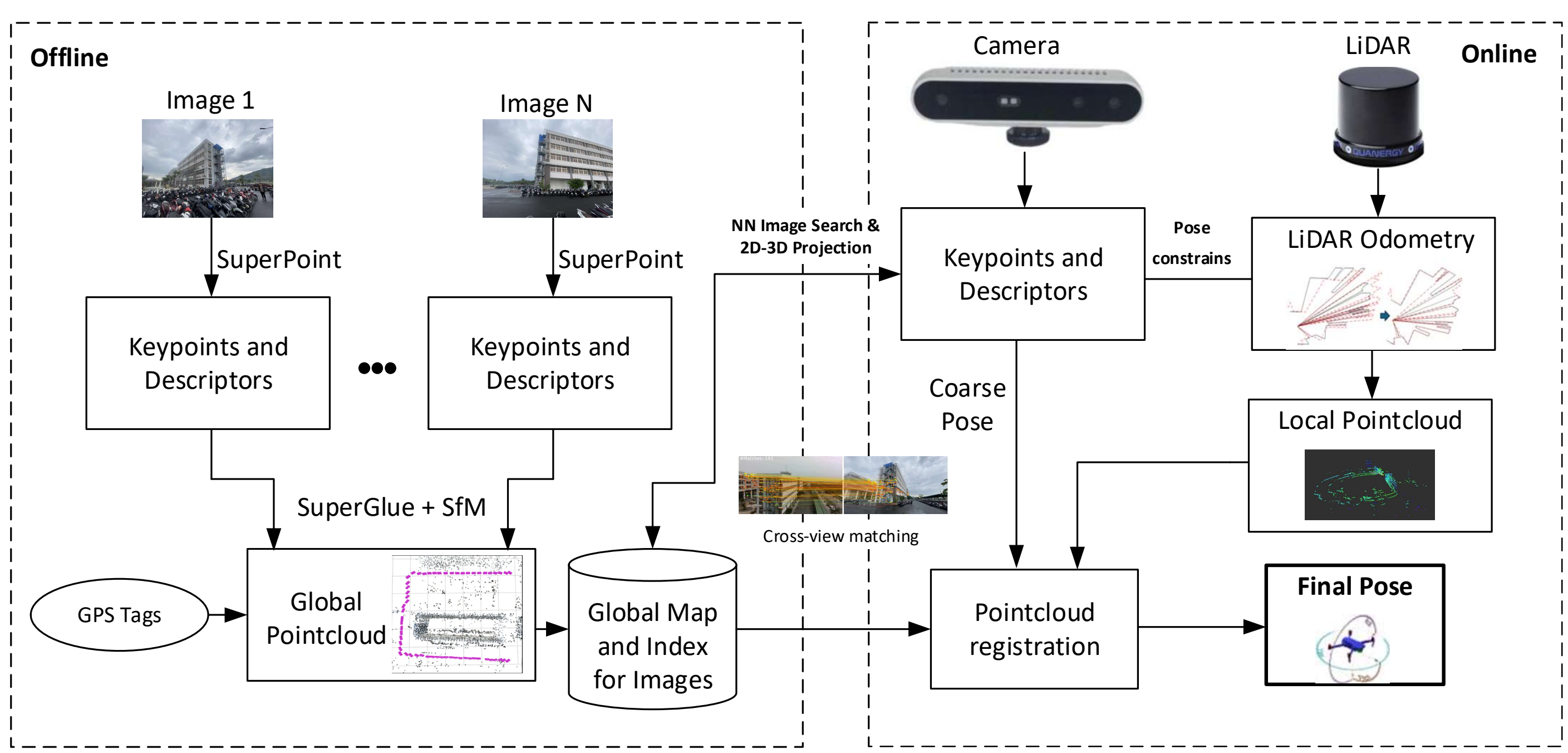

Figure 1: Proposed localization method

Furthermore, during low-altitude infrastructure inspections, lighting conditions can vary across sessions, while environments such as bridge undersides often contain unevenly distributed visual features [6, 7]. As a result, sensor data collected from a single flight may be insufficient for consistent localization. To address this limitation, prior data gathered across multiple sessions can be exploited. In structural health monitoring applications, for example, a baseline survey is often established using ground agents or manual flights. Our approach leverages this pre-existing visual database and integrates complementary cross-session information to enrich the feature space, correct drift, and provide the robustness required for UAVs to localize reliably.

In this work, we propose Cross-Fusion, a hybrid localization method that integrates LiDAR and camera data across multiple sessions through a deep learning framework. While individual components such as feature extraction, global retrieval, and LiDAR odometry draw on established methods, the novelty of Cross-Fusion lies in the following contributions:

- Cross-session fusion strategy: A novel pipeline that associates online UAV observations with a pre-built visual-geometric database constructed from prior inspection sessions, such as ground agents or manual flights. This cross-session association provides global pose correction constraints that reduce drift and improve robustness in feature-sparse environments.

- LiDAR-guided 2D-3D correspondence transfer: A new mechanism that lifts 2D-2D image feature matches into 3D-3D point-cloud correspondences by projecting matched keypoints through the online LiDAR stream for 6 Degrees-of-Freedom (DoF) pose correction without an IMU or stereo camera.
- Robust and lightweight sensor setup: The above contributions are realized with only a 3D LiDAR and a monocular camera, avoiding the complexity of tightly coupled visual-inertial systems while maintaining competitive accuracy and greater robustness under low frame rates or degraded visual conditions.
- Open-source implementation: The implementation is made available to support reproducibility and future research at: https://github.com/quachconghoang/uav_lidar_hloc.

## 2. Related work

In the literature, sensor-based localization for UAVs can be broadly grouped into three main categories: Simultaneous Localization and Mapping (SLAM)/odometry-based methods, reference-map or absolute visual localization methods, and learning-based methods [8, 9, 10]. SLAM and odometry approaches estimate the UAV pose from onboard sensor observations using cameras, LiDAR, IMU, or their combinations. Visual SLAM builds a map of the environment while estimating the camera pose [11]. It typically uses front-end visual odometry to estimate motion between frames and back-end optimization (loop closure) to reduce accumulated drift.

The front-end visual odometry (VO) estimates relative motion by tracking image features or minimizing photometric error between consecutive frames. Although VO is more lightweight than full SLAM, its recursive nature means that drift can accumulate when loop closure or global correction is unavailable [8, 11]. Recent VO studies show that feature correspondences with low reprojection error still introduce orientation bias, which can reduce long-term pose accuracy. For example, stereo orientation-prior VO uses stereo geometry and sum-of-squares feasibility constraints to reject misleading correspondences, but it still cannot eliminate drift [12]. This issue is especially challenging for UAVs because they must estimate full 6-DoF motion including altitude, pitch, and roll under strict size, weight, and power constraints [9].

LiDAR-based SLAM provides an alternative by using geometric information from point clouds. Compared with camera-only methods, LiDAR SLAM is less sensitive to illumination changes and can provide more reliable localization in visually degraded environments [13, 14]. However, LiDAR-only methods can still fail in geometrically degenerate scenes such as long tunnels, open spaces, or areas dominated by planar surfaces. Typical scan-to-scan or scan-to-map matching methods also accumulate error over time and require loop closure to maintain global consistency.

To exploit the complementary strengths of geometry and appearance, recent research has focused on LiDAR-visual-inertial localization. Visual-LiDAR Odometry and Mapping (V-LOAM) combines high-frequency visual odometry with lower-frequency LiDAR odometry, where the visual module provides rapid motion estimates and the LiDAR module refines the trajectory and corrects drift under fast motion and temporary visual degradation [15]. LiDAR-Visual-Inertial Smoothing and Mapping (LVI-SAM) further integrates LiDAR, camera, and IMU measurements in a factor-graph framework [16]. It utilizes LiDAR depth to improve feature tracking and visual estimates to support LiDAR scan matching and loop closure. Continuous-time Laser-Visual-Inertial (Ct-LVI) odometry and mapping extends this direction by formulating the trajectory in continuous time and using probabilistic submaps to enable tighter fusion of asynchronous laser, visual, and inertial measurements in large-scale or fast-motion scenarios [17]. Other tightly coupled systems such as Robust, Real-time, LiDAR-Inertial-Visual Estimator and Mapping (R2LIVE) combine LiDAR, inertial, and visual measurements within an iterated Kalman filter and factor-graph optimization to improve robustness under aggressive motion, sensor failure, and tunnel-like environments [18]. Fast LiDAR-Inertial-Visual Odometry (FAST-LIVO) improves computational efficiency by using a sparse-direct formulation: raw LiDAR points are registered to an incremental point-cloud map, while image patches attached to map points are used for direct photometric alignment without explicit visual feature extraction [19]. These systems demonstrate strong real-time performance and robustness, but they are mainly designed for online, single-session odometry or SLAM. They do not address how prior data from multiple inspection sessions can be reused to improve localization in long-term or cross-session infrastructure inspection.

Learning-based localization has also become an active research direction. Deep neural networks can learn complex non-linear relationships between sensor measurements and pose-related quantities to provide an alternative to geometric pipelines. Convolutional Neural Network (CNN)-based and multi-task learning models have been used to estimate monocular depth, optical flow, and ego-motion for UAV navigation in dynamic environments [20]. Sequence models have also been explored for navigation during Global Navigation Satellite System (GNSS) outages. For example, Recurrent Neural Network/Long Short-Term Memory (RNN/LSTM)-based sensor fusion has been used to estimate UAV position from alternative navigation sources such as holographic radar and radio positioning systems [21]. In broader GPS-outage vehicle localization, cascaded learning models combine motion-state classification, temporal modeling, and self-attention to predict pseudo-GPS positions under different motion patterns [22]. More recently, self-attention LSTM networks have been proposed for UAV integrated navigation in GNSS-denied environments, where learned attitude and velocity estimates are fused into an Extended Kalman Filter (EKF) framework to reduce long-term position drift [23].

Despite their promise, learning-based methods usually require large and representative training datasets, careful parameter tuning, and substantial computational resources for real-time inference. Their performance may also degrade under domain shifts caused by changes in viewpoint, lighting, weather, scene appearance, or vehicle dynamics. In contrast, classical SLAM and odometry pipelines are more interpretable but remain sensitive to drift, visual texture loss, LiDAR degeneracy, and the absence of reliable loop closure. These limitations motivate the use of prior multi-session information for UAV localization in infrastructure inspection.

To address these limitations, this work proposes a cross-session SLAM approach for UAV localization in GPS-denied infrastructure inspection. Unlike conventional single-session SLAM or odometry systems, the proposed approach utilizes a baseline visual-geometric database constructed from prior inspection sessions. During online operation, LiDAR odometry provides continuous local motion estimation, while visual retrieval and cross-session matching associate the current UAV observations with the baseline database. These cross-session associations provide global constraints for pose correction which help to reduce drift in visually degraded or geometrically ambiguous regions. By combining online LiDAR odometry with prior multi-session visual-geometric information, the proposed approach improves localization robustness without relying on GPS or large task-specific training datasets.

## 3. Methodology

Inspired by deep learning, the Cross-Fusion framework is designed with two phases, offline and online, to integrate data collected across different sessions, as illustrated in Figure 1. In describing our methodology, we distinguish between algorithms adopted from the literature and the integration strategy proposed in this study. The individual modules for feature extraction, global description, structure-from-motion (SfM), and point-cloud registration are based on established methods. Our main contribution is the pipeline that connects these modules, particularly the mechanism that links the online 3D LiDAR odometry stream with the offline visual 3D map through global descriptor retrieval and 2D-3D feature mapping to correct localization drift.

### 3.1. Offline Phase

In the offline phase, an image database of the operating environment is constructed. For routine inspection scenarios, this baseline data is collated using cameras on mobile platforms, ground robots, or manually piloted drones during an initial survey. Each image in the database is then processed to extract local features. To ensure high performance, CNN-based keypoint detectors and descriptors such as SuperPoint and DIScrete Keypoints (DISK) are employed [24][25]. As a result, each image is represented by a set of keypoints $\mathbf{u}_i$ along with their associated descriptors. These local descriptors are then used to generate a global descriptor for each image using methods such as EigenPlaces [26]. The use of global descriptors is critical for system scalability as it allows the online system to rapidly retrieve candidate images from large-scale databases without performing exhaustive local feature matching. Next, a sparse 3D reference model of the environment is built from the database images. To achieve it, images covering overlapping areas such as images taken at consecutive time are first identified. Each pair of overlapping images is then matched based on their local feature descriptors using matchers such as SuperGlue [25]. Finally, all matched pairs are fed into a SfM pipeline such as COLMAP [27] to triangulate 3D points and estimate camera poses. The outcome is a sparse map of 3D points $\mathbf{X}_i$ with known camera positions.

### 3.2. Online Phase

In the online phase, the UAV's pose is estimated using two data streams: LiDAR and camera. LiDAR provides a coarse estimate of the pose through an odometry technique. In feature-sparse zones, the system relies more heavily on this continuous LiDAR stream. This estimate is then refined using visual data from the camera and the global sparse map generated during the offline phase.

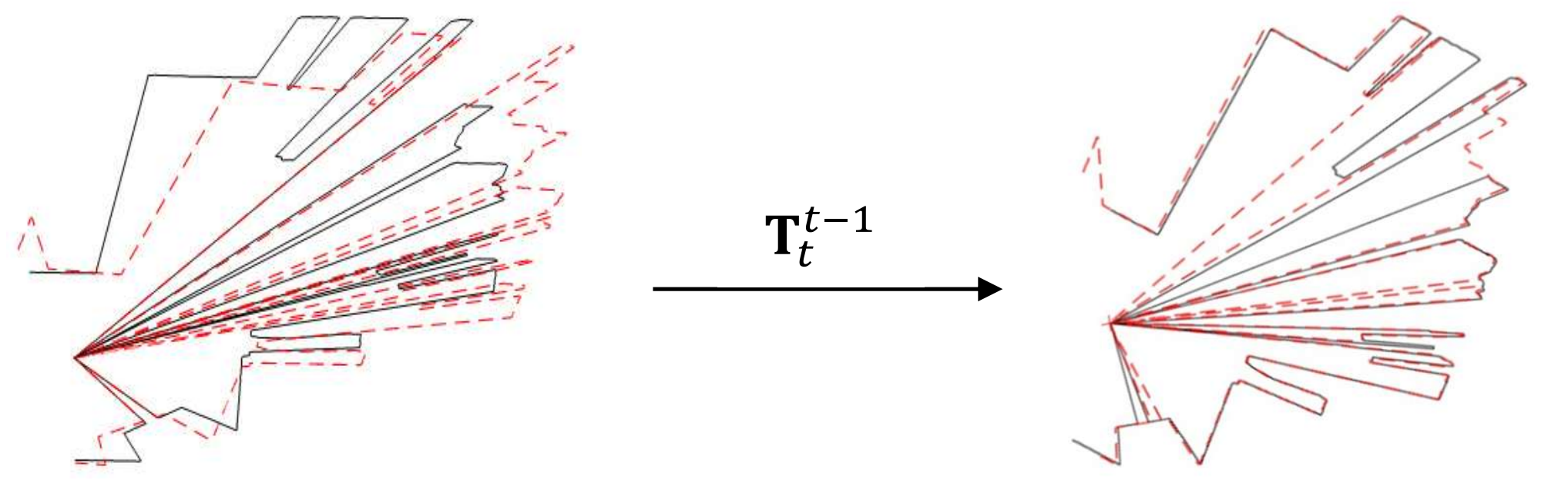

a) LiDAR scans before alignment b) LiDAR scans after alignment

Figure 2: Illustration of the LiDAR scans alignment

#### 3.2.1. Pose Estimation Using LiDAR Odometry

LiDAR odometry estimates the relative motion between consecutive LiDAR scans over time, as illustrated in Figure 2. At each timestep $t$, the UAV receives a point cloud $\mathbf{L}_t$ from the LiDAR sensor:

$$\mathbf{L}_t = \left\{\boldsymbol{p}_i^t \in \mathbf{R}^3\right\}_{i=1}^{N_t} \tag{1}$$

It is used to compute the transformation $\mathbf{T}_{t-1}^{t}$ that aligns it with the previous scan $\mathbf{L}_{t-1}$:

$$\mathbf{T}_t^{t-1} = \begin{bmatrix} \mathbf{R}_t & \mathbf{t}_t \\ 0 & 1 \end{bmatrix}, \tag{2}$$

where $\mathbf{R}_t$ and $\mathbf{t}_t$ are the rotation and translation matrices at time $t$, respectively. The transformation matrix representing the UAV pose, $\mathbf{P}_t$, relative to its initial pose, $\mathbf{P}_0$, then can be computed as:

$$\mathbf{P}_t = \mathbf{T}_1^0 \mathbf{T}_2^1 \ldots \mathbf{T}_t^{t-1} \mathbf{P}_0 \tag{3}$$

One common method to estimate $\mathbf{T}_t^{t-1}$ is Iterative Closest Point (ICP) [28], which solves for the transformation that minimizes the distance between corresponding points in two scans:

$$\min_{\mathbf{R},\mathbf{t}} \sum_i \left\| \boldsymbol{p}_i^t - (\mathbf{R}\boldsymbol{p}_i^{t-1} + \mathbf{t}) \right\|^2 \tag{4}$$

Apart from ICP, other methods such as Cartographer [29], Fast LiDAR-Inertial Odometry (Fast-LIO2) [30], or GLIM [31] can also be used to estimate $\mathbf{T}_t^{t-1}$. The advantage of LiDAR odometry is fast processing time and update rate. However, using LiDAR data alone subject to drift because it works by comparing consecutive scans. The error after each alignment is not correct and accumulated over time resulting in significant drift. To address this issue, our method employs visual data as a means to correct the drift.

#### 3.2.2. Pose Correction Using Visual Data

For an incoming image captured from UAV in the online phase, its location is first identified by comparing its similarity with database images. In particular, a global image descriptor is computed for each input image and then compared with the database images to find the k most similar (top-k) matches. This coarse retrieval step significantly reduces search space and provides the initial location hypotheses, which is crucial for maintaining real-time operation as the size of the baseline database grows. Once a set of candidate database images is retrieved, fine-grained feature matching is performed between the input image and each candidate. Local keypoints in the input image is extracted and matched to keypoints in the candidate database images. Again, SuperGlue can be used to compute 2D-2D correspondences between the input and each database image. These 2D-2D matches are transferred to 3D-3D correspondences via the LiDAR data: the LiDAR point cloud $\mathbf{L}_i$ collected at the time taken the input image is compared with 3D points $\mathbf{X}_i$ from the map having the camera pose for the matched image. Thus, we obtain a set of correspondences $\{(\mathbf{L}_i, \mathbf{X}_i)\}$ between input LiDAR points $\mathbf{L}_i$ and the point cloud generated in the offline phase $\mathbf{X}_i$. A truncated point cloud registration problem is then solved to recover the camera pose $(\mathbf{R}, \mathbf{t})$ that aligns the two point clouds. Specifically, $(\mathbf{R}, \mathbf{t})$ is computed by minimizing the distance error:

$$\min_{\mathbf{R},\mathbf{t}} \sum_i \|\mathbf{L}_i - (\mathbf{R}\mathbf{X}_i + \mathbf{t})\|^2 \tag{5}$$

Finally, the optimal $(\mathbf{R}, \mathbf{t})$ provides the UAV's 6-DoF pose relative to the global map.

### 3.3. Pseudocode of the algorithm

The pseudocode of the proposed algorithm is presented as follows:

```
# ==================================================================================
# Offline Phase
# ==================================================================================
function OfflinePhase():
    # Step 1: Build image database
    images = collectImages(devices=['camera', 'mobile', 'drone'])

    # Step 2: Extract local features (CNN-based) → descriptors
    for image in images:
      keypoints, descriptors = extractLocalFeatures(image,method='SuperPoint/DISK') # [19],[21]
      imageDB.store(image, keypoints, descriptors)

    # Step 3: Compute global descriptor for each image
    for image in imageDB:
      globalDesc = computeGlobalDescriptor(image, method='EigenPlaces')             # [20]
      imageDB.storeGlobalDescriptor(image, globalDesc)

    # Step 4: Build sparse 3D map
    overlappingPairs = findOverlappingImages(imageDB)
    matchedPairs = matchImagePairs(overlappingPairs, matcher='SuperGlue')           # [21]
    sparseMap, cameraPoses = runSfM(matchedPairs, tool='COLMAP')                    # [22]
    return sparseMap, imageDB

# ==================================================================================
# Online Phase
# ==================================================================================
function OnlinePhase(lidarStream, cameraStream, sparseMap, imageDB):
    for each time t:
        # Step 1: Estimate coarse pose using LiDAR odometry
        P_t = lidarStream.getPointCloud(t)                                          # Eq. (1)
        T_rel = computeTransformation(P_t, P_t-1)                                   # Eq. (2)
        T_t = composePose(T_t-1, T_rel)                                             # Eq. (3)
        # Solve via ICP
        T_t = runLiDAROdometry(P_t, method='ICP')                                   # Eq. (4)

        # Step 2: Correct drift using visual data
        I_t = cameraStream.getImage(t)

        # Step 2.1: Coarse retrieval based on global descriptors
        desc_t = computeGlobalDescriptor(I_t)
        candidates = retrieveTopK(desc_t, imageDB)

        # Step 2.2: Fine-grained 2D-2D feature matching with candidates
        for candidate in candidates:
```

```
        kp_input, kp_db = matchKeypoints(I_t, candidate, method='SuperGlue')

        # Step 2.3: Establish 3D-3D correspondences via LiDAR and map
        P_lidar = lidarStream.getPointCloud(t)
        P_map = sparseMap.get3DPoints(candidate)
        correspondences = find3DCorrespondences(kp_input, kp_db, P_lidar, P_map)

        # Step 2.4: Align point clouds using ICP to refine the pose            # Eq. (5)
        T_refined = min_distance_ICP(correspondences)

        if isValidPose(T_refined):
            T_t = T_refined
            break

    outputPose(T_t)
```

## 4. Results

A number of experiments have been conducted to evaluate the validity of the proposed method as follows.

### 4.1. Experimental setup

The hardware platform consists of a quadcopter equipped with a LiDAR, a camera, and an IMU for localization, as shown in Figure 3. The LiDAR is a Quanergy M8 [32], which provides 8-layer scanning, a range of 100 meters, an update rate of 20 Hz, a 360° horizontal field of view, and a 20° vertical field of view. It outputs point cloud data for LiDAR odometry and position correction. The camera is an Intel RealSense D435, capturing color images at a resolution of 1920 × 1080 pixels with a horizontal and vertical field of view of 69° × 42° at 30 frames per second. These images are the primary source for extracting visual features of the environment.

Experiments are conducted in the surrounding area of a building with the size of 150 m x 150 m. In the offline phase, 60 images are collected from a smartphone at the resolution of 4032 x 3024 to generate the database image. In the online phase, data collected by the LiDAR and camera on the UAV is processed to estimate the UAV positions. Figure 4 shows the visual and point cloud data collected from the camera and LiDAR during experiments. During that time, GPS data is collected, not for the localization but as a reference to evaluate the accuracy of the proposed method.

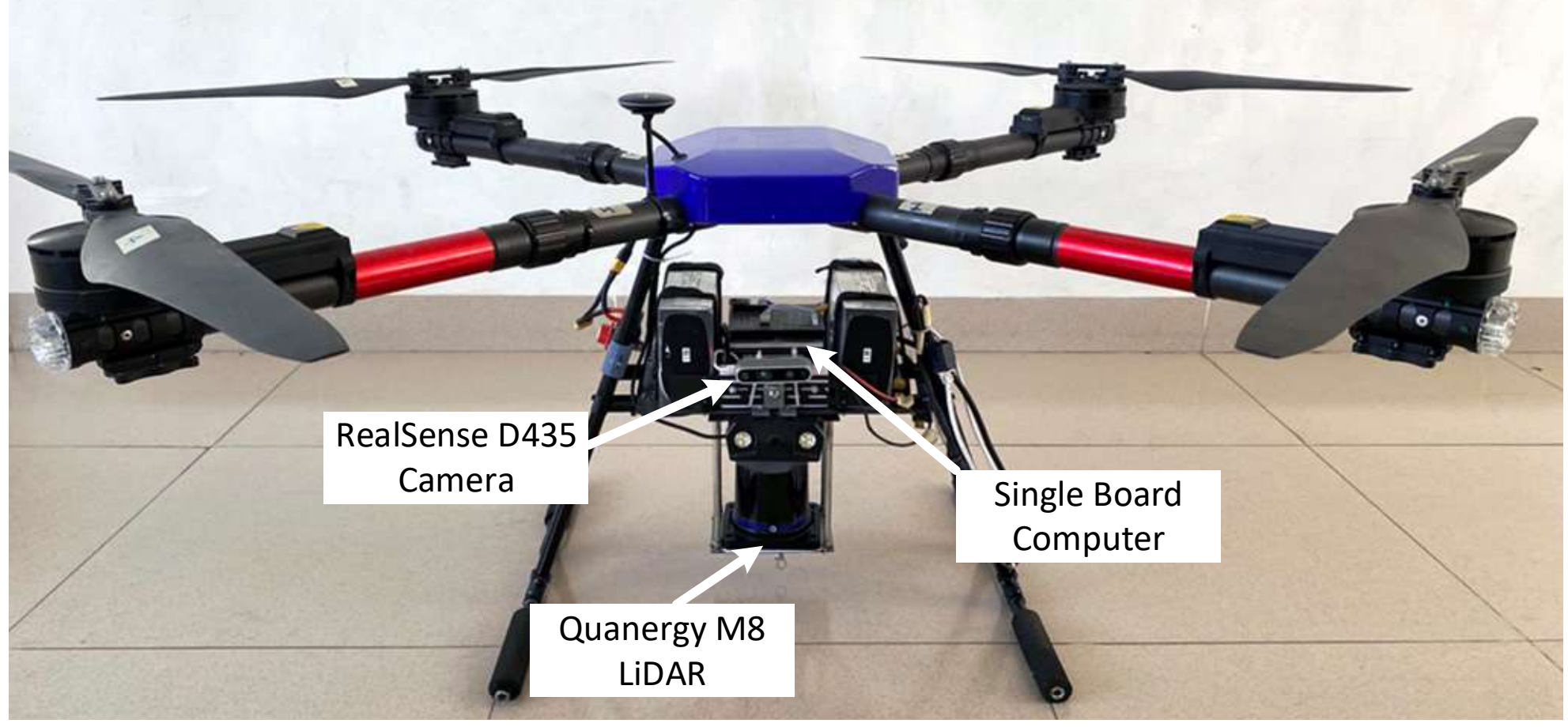

Figure 3: The quadcopter drone with sensors and a single board computer for experiments

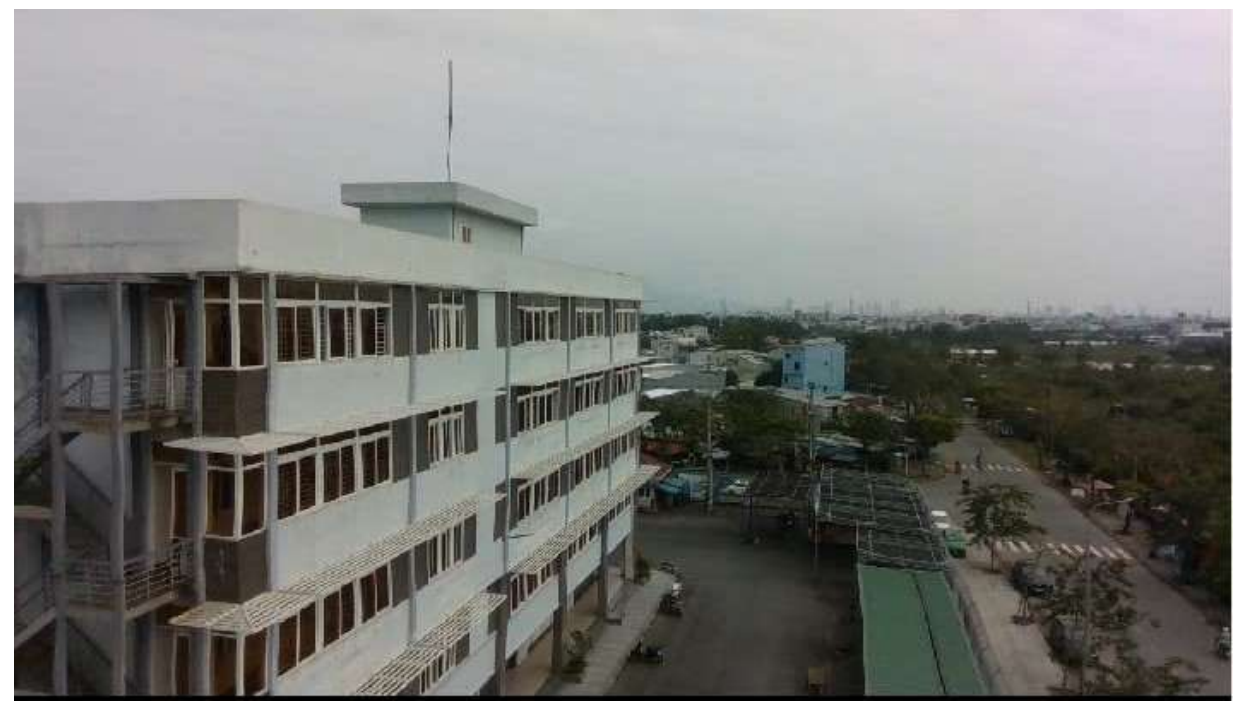

a) Visual Data from Camera

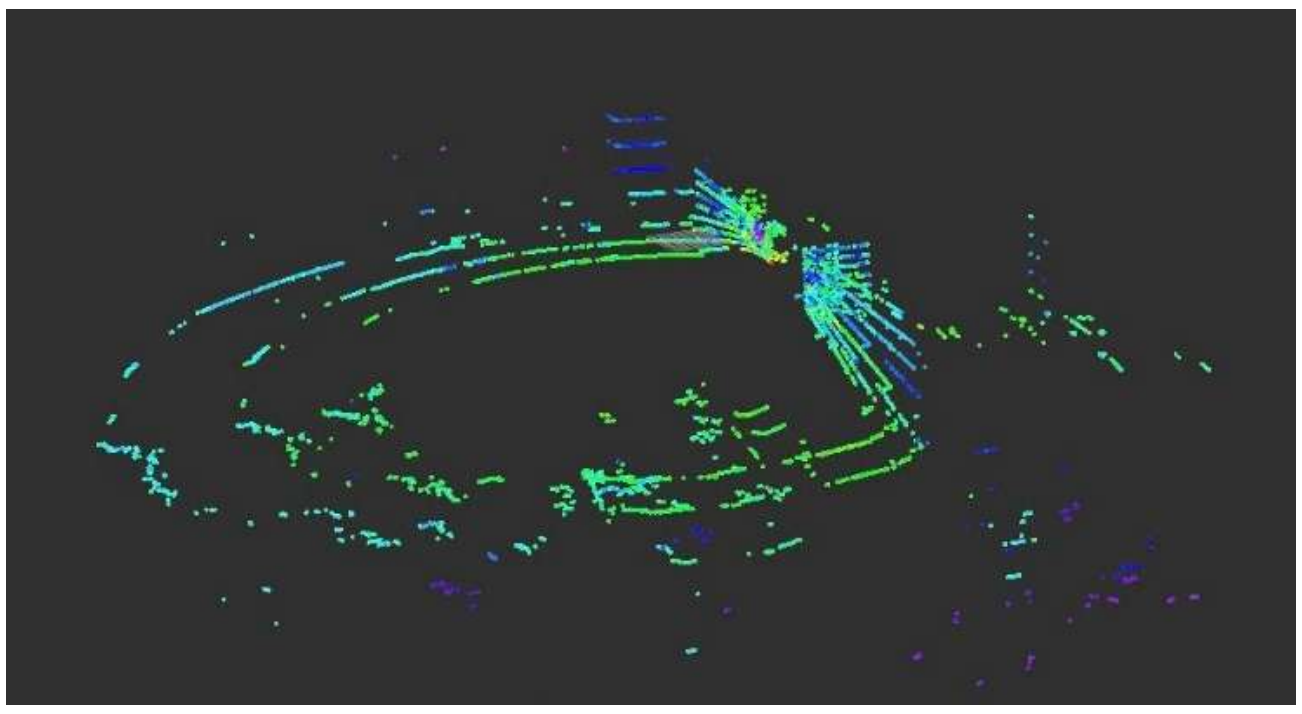

b) Point cloud Data from LiDAR

Figure 4: Data collected from the camera and LiDAR

### 4.2. Experimental Results

Figure 5 shows the estimated camera poses (in purple) and point cloud data (in gray) reconstructed from the image database. The camera poses appear to be well aligned with minimal errors. In contrast, the point cloud is relatively noisy due to the limitations of generating 3D structure from 2D images using the Structure-from-Motion (SfM) method. Despite the noise, the reconstructed poses and point cloud are sufficient for fusion with image and LiDAR data in the online localization phase, as illustrated in Figure 6. The UAV trajectory estimated by our method aligns more closely with the GPS reference than the one produced by LiDAR-based odometry. Notably, our method maintains stable localization without the drift observed in the LiDAR-only approach.

This observation is further supported by Figure 7, which shows the localization error along selected segments of the trajectory. The LiDAR-based method accumulates drift over time, leading to a larger error than the proposed fusion method. In contrast, our method maintains a lower estimation error by using visual corrections from the offline database. The slight decrease in the LiDAR error near the end should not be interpreted as drift recovery. Rather, it occurs because the drifted LiDAR trajectory becomes temporarily closer to the GPS reference in that segment, while the accumulated drift itself remains. Finally, Figure 8 demonstrates successful feature matching between an online input image and a corresponding image from the database. The features are accurately matched which enables reliable location recognition for effective localization.

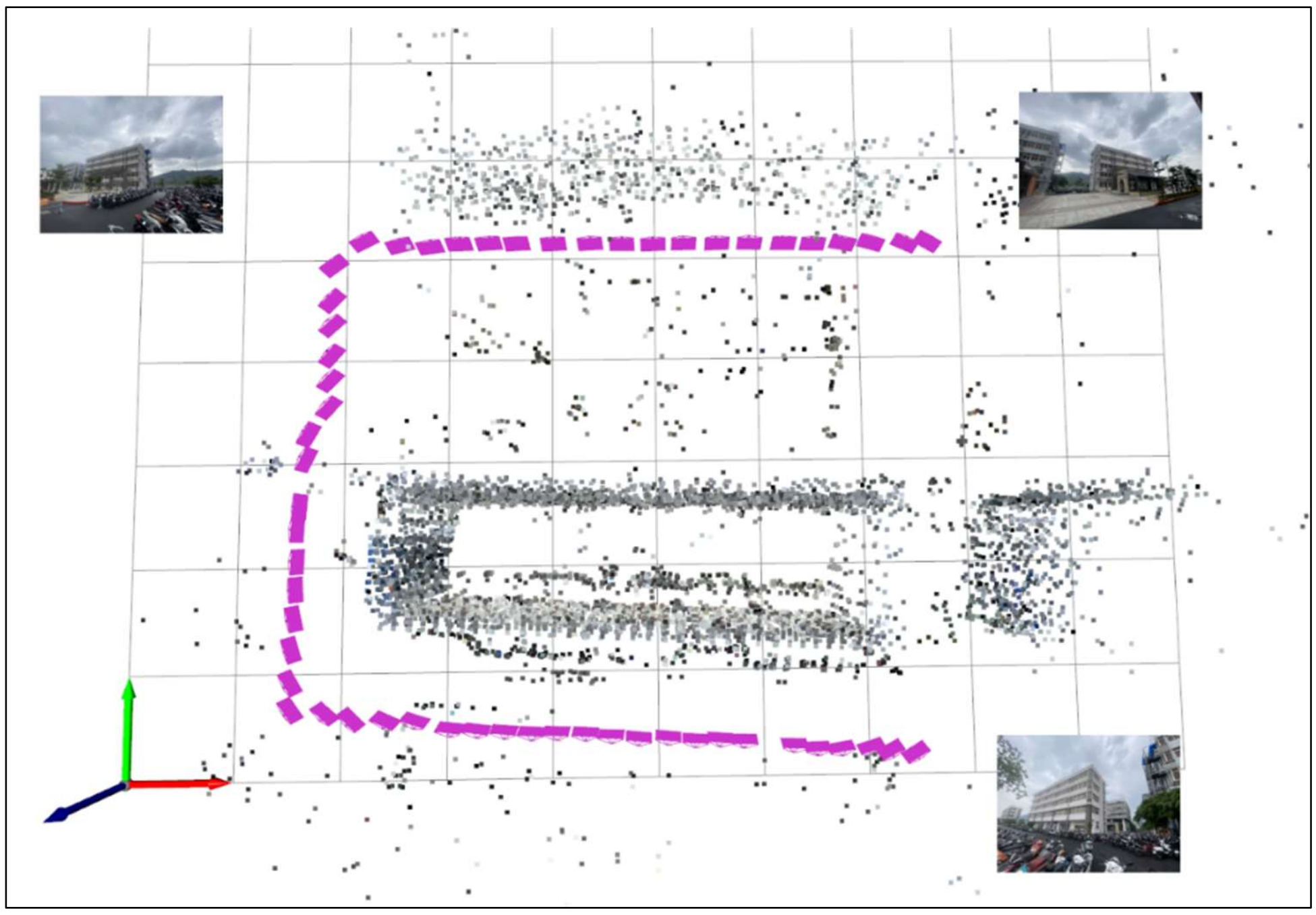

Figure 5: Camera poses (purple) and point cloud data (gray) estimated from the image database in the offline phase.

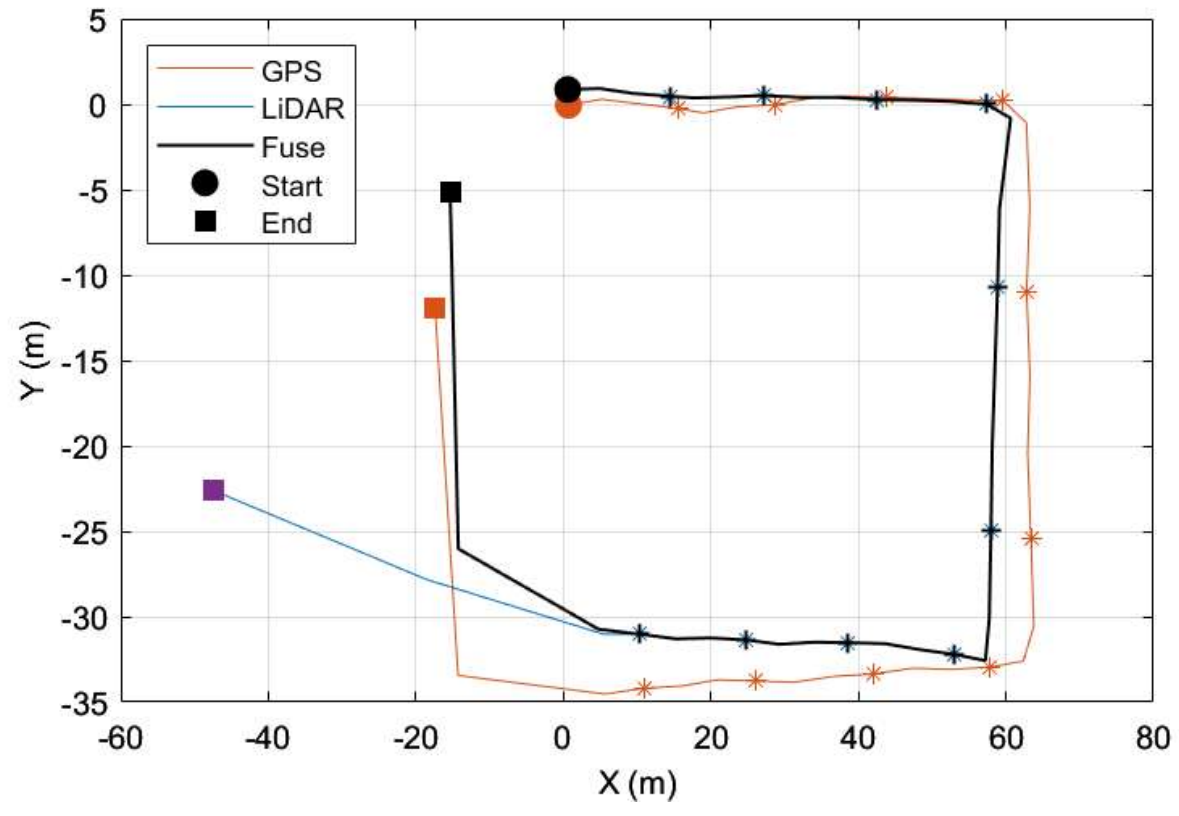

Figure 6: Trajectories from LiDAR odometry (blue), GPS (orange), and our method (black).

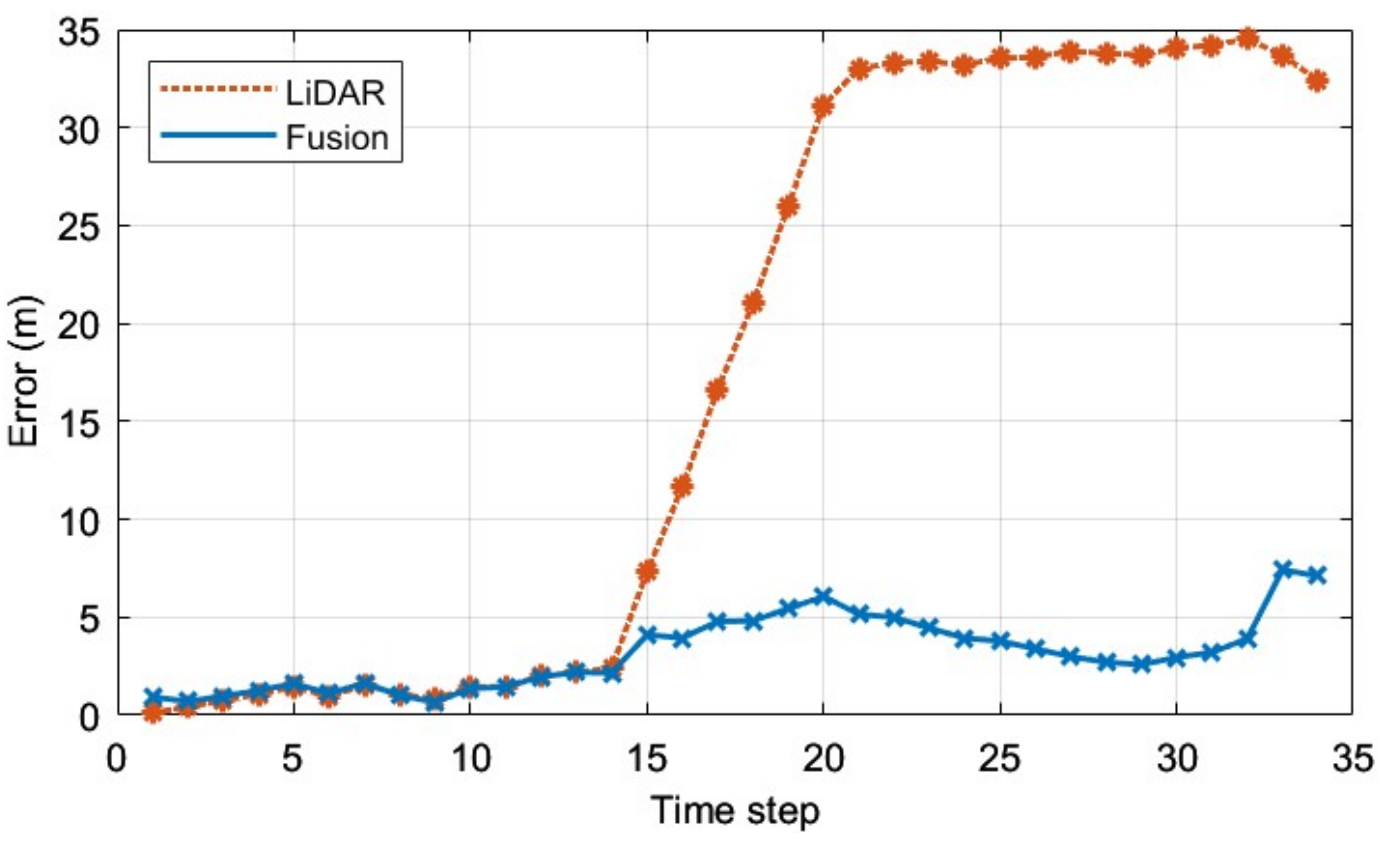

Figure 7: Localization error of LiDAR odometry and the proposed fusion method at keyframes sampled every 10s.

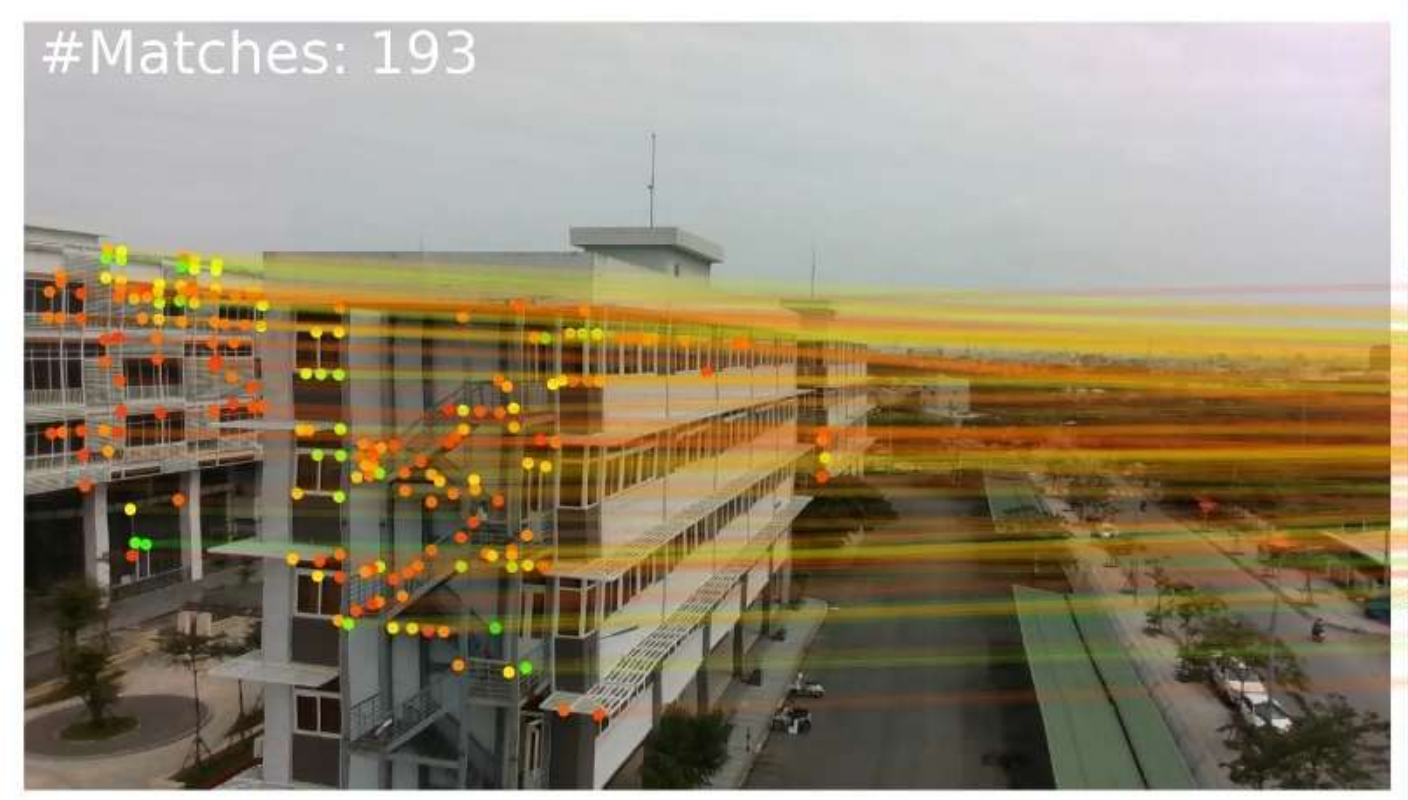

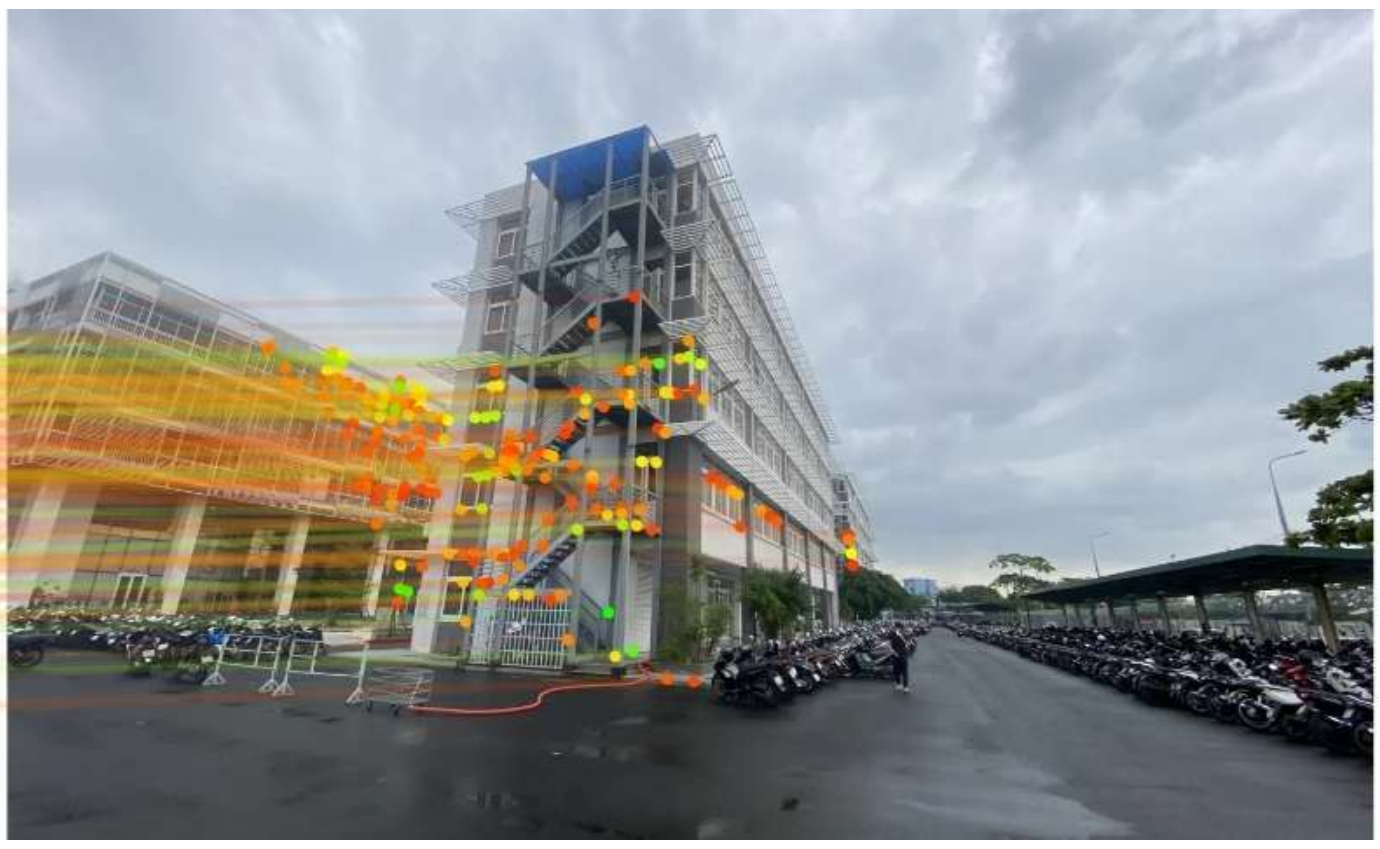

Figure 8: Feature matching between the images collected online and offline

In a second experiment, we compared our algorithm with FAST-LIVO [19], a state-of-the-art SLAM method for UAVs. Under normal operating conditions, FAST-LIVO achieves higher accuracy than our approach, as illustrated in Figures 9 and 10. This outcome is expected as FAST-LIVO is a tightly coupled LiDAR-Inertial-Visual (LIVO) SLAM framework that jointly optimizes all sensor measurements at the raw/feature level, whereas our method adopts a loosely coupled fusion strategy. Nevertheless, our method demonstrates greater robustness in extreme scenarios, such as when the camera frame rate is very low or when the drone flies at high speed leading to limited overlap between consecutive frames. Figure 11 shows the trajectory estimated by FAST-LIVO when the frame rate is deliberately reduced to below 1 fps. The resulting divergence indicates that FAST-LIVO fails to estimate the trajectory under these harsh conditions due to its tightly coupled design. In this regard, our proposed method provides a balanced trade-off between accuracy and robustness.

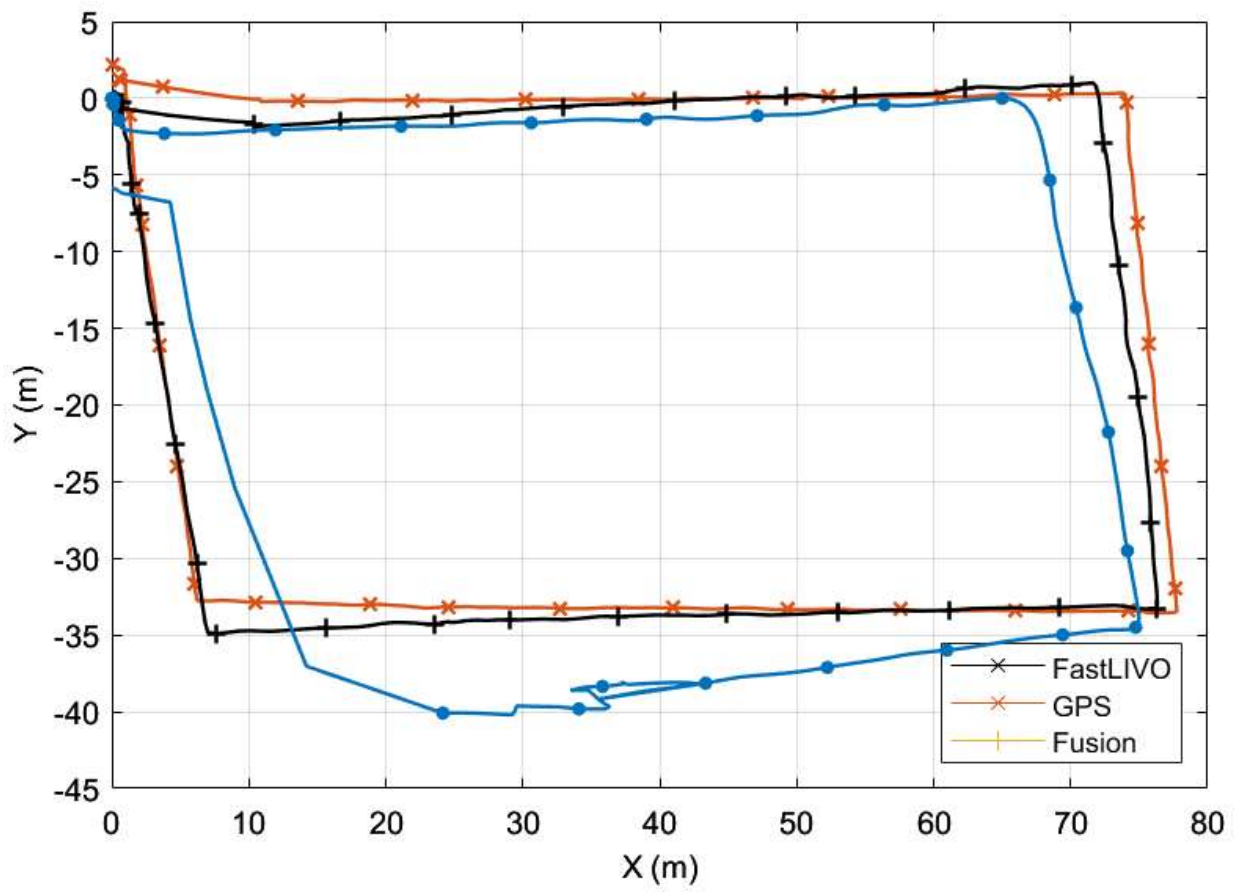

Figure 9: Trajectories from FAST-LIVO (black), GPS (orange), and our method (blue).

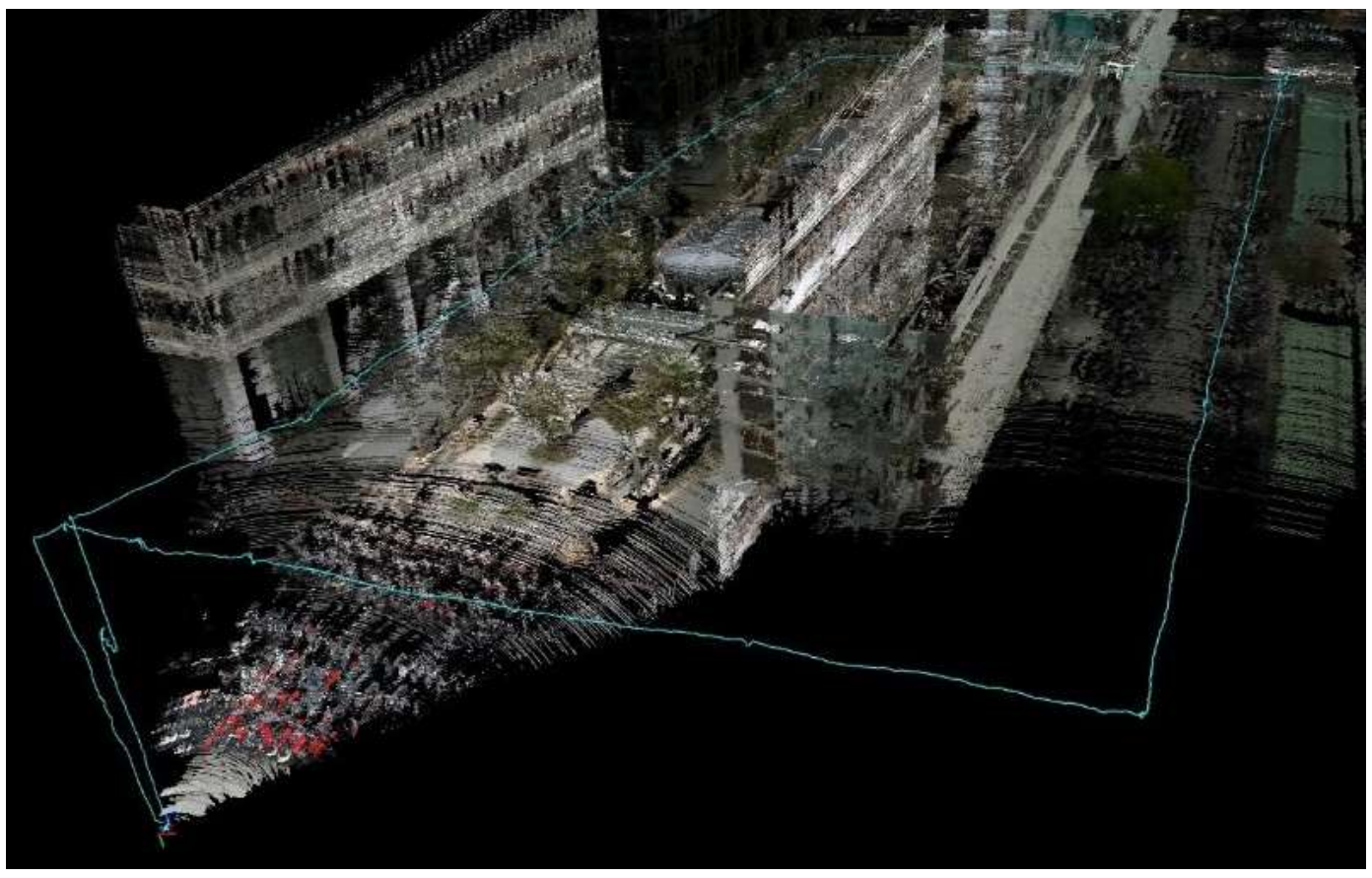

Figure 10: Trajectory estimated by FAST-LIVO under normal conditions.

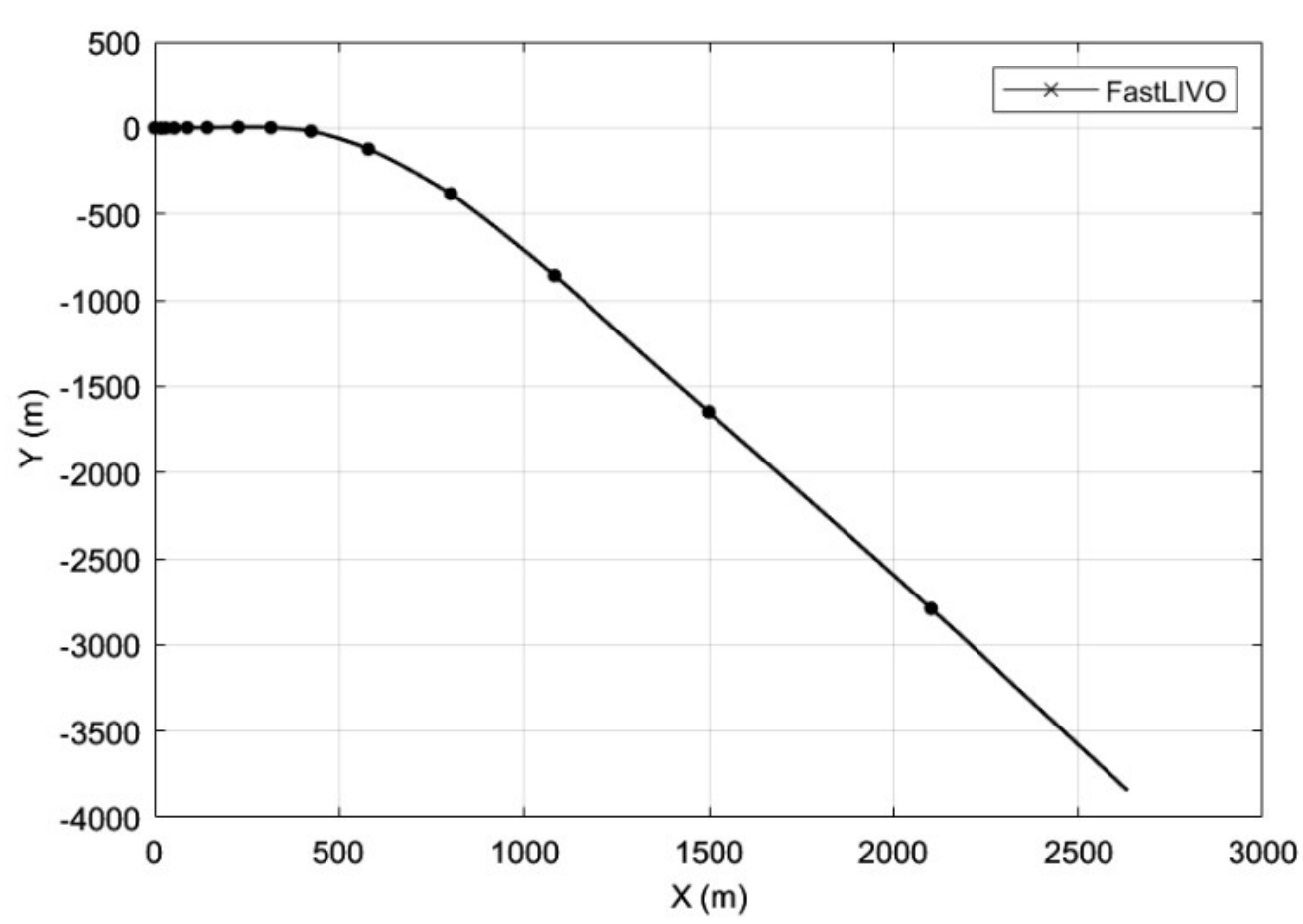

Figure 11: FAST-LIVO trajectory estimation failure at 0.5 fps.

## 5. Conclusion

In this work, we have introduced Cross-Fusion, a robust and practical solution for UAV localization designed for low-altitude economy applications, such as routine infrastructure inspection in GPS-denied environments. By fusing spatial data from a 3D LiDAR with visual information from a monocular camera across different sessions, Cross-Fusion achieves a balance between reliable localization and accuracy. The system leverages baseline structural surveys to build its offline database and mitigates drift in feature-sparse areas by using LiDAR odometry supported by global descriptor retrieval. The system requires a simple hardware setup and is made openly available to facilitate reproducibility and support further research.

Despite its strengths, Cross-Fusion currently depends on a pre-collected image database for online matching, which limits its applicability in dynamic scenarios where pre-mapping is impractical. Future work will address this limitation by integrating data from multiple online sessions such as coordinating ground and UAV agents to enable real-time map construction and localization without relying solely on prior static data collection.